\documentclass{article}

\usepackage{arxiv}

\usepackage[utf8]{inputenc} 
\usepackage[T1]{fontenc}    
\usepackage{hyperref}       
\usepackage{url}            
\usepackage{booktabs}       
\usepackage{amsfonts}       
\usepackage{nicefrac}       
\usepackage{microtype}      
\usepackage{cleveref}       
\usepackage{lipsum}         
\usepackage{graphicx}
\usepackage[numbers]{natbib}
\usepackage{doi}

\usepackage{subcaption}

\title{Federated Learning for Cross-Regional Epidemic Forecasting with Asynchronous Outbreaks}

\date{}

\usepackage{authblk}

\newbox{\orcid}\sbox{\orcid}{\includegraphics[scale=0.06]{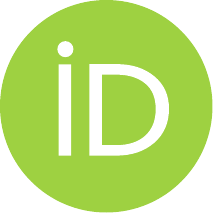}}
\author[1,2]{%
    \href{https://orcid.org/0000-0002-8252-5585}{\usebox{\orcid}\hspace{1mm}Ruibin Liu}%
}
\author[3]{Francisco Fernandes}
\author[1,2]{Xuanfeng Li}
\author[1,2]{Zige Liu}
\author[3]{Andr\'e Salgado}
\author[1,2]{Chitin Hon\thanks{\texttt{cthon@must.edu.mo}}}
\author[4,5]{Zifeng Yang\thanks{\texttt{jeffyah@163.com}}}
\author[1,3]{Arlindo Oliveira\thanks{\texttt{arlindo.oliveira@tecnico.ulisboa.pt}}}
\author[3]{Ana Teresa Freitas\thanks{\texttt{ana.freitas@tecnico.ulisboa.pt}}}

\affil[1]{Respiratory Disease AI Laboratory in Epidemic Intelligence and Applications of Medical Big Data Instruments, Faculty of Innovation Engineering, Macau University of Science and Technology, Taipa, Macau SAR, China.}
\affil[2]{Institute of Systems Engineering, Macau University of Science and Technology, Taipa, Macau SAR, China.}
\affil[3]{INESC-ID, Instituto Superior Técnico, University of Lisbon, 1649-004 Lisbon, Portugal}
\affil[4]{National Clinical Research Center for Respiratory Disease, State Key Laboratory of Respiratory Disease, Guangzhou Institute of Respiratory Health, the First affiliated Hospital of Guangzhou Medical University, Guangzhou Medical University, Guangzhou, Guangdong, 510120, P.R. China.}

\renewcommand{\shorttitle}{}

\hypersetup{
pdftitle={Federated learning for epidemic forecast},
pdfsubject={cs.LG, stat.ML, q-bio.QM},
pdfauthor={Ruibin Liu},
pdfkeywords={Federated Learning, Long Short-Term Memory, Cross-border knowledge exchange, COVID-19 Spread Forecast, Infectious disease forecast},
}

\begin{document}
\maketitle

\begin{abstract}
\textbf{Objective}
Epidemic outbreaks are often asynchronous across regions, yet collaborative forecasting is hampered by privacy laws and heterogeneous public health interventions. We developed and evaluated, as a proof of concept, a federated learning (FL) framework for cross-regional COVID-19 forecasting under asynchronous outbreaks among Guangdong, Macau, and Portugal, while maintaining data sovereignty.

\textbf{Methods}
Using daily case data from WHO and China CDC for the three regions—whose outbreaks were asynchronous and interventions heterogeneous—we trained Long Short-Term Memory networks via FL for short-term forecasting, benchmarked against localized and centralized training using Root Mean Square Error, Mean Absolute Error, and Symmetric Mean Absolute Percentage Error. All experiments used the Flower FL framework with consistent hyperparameters across sites. Publicly available, aggregated case-count data were used so that the centralized benchmark could be computed as a reference.

\textbf{Results}
The dataset comprised 1,233 daily observations (411 days) of new confirmed COVID-19 cases from November 24, 2021, to January 8, 2023. The earlier Omicron wave in Portugal supplied transmission dynamics that improved forecasts in the later-onset regions of Macau and Guangdong. The FL model achieved a normalized Root Mean Square Error of 0.028, an 88.3\% improvement over the local model (0.240) and comparable to the centralized model, with network overhead mainly determined by model size and aggregation rounds.

\textbf{Conclusion}
FL achieved short-term forecasting accuracy close to centralized training while maintaining data residency, with acceptable overhead. Leveraging asynchronous outbreaks across regions may offer a practical path to privacy-preserving collaborative forecasting, though prospective validation is required.
\end{abstract}


\section{Introduction}
In recent years, the COVID-19 pandemic and previous outbreaks like Ebola have revealed critical gaps in global disease surveillance systems, particularly the inability to rapidly and securely share data across jurisdictions \cite{hill2024realising}. Data privacy, national data sovereignty, and technical heterogeneity remain major barriers to effective, timely global collaboration \cite{fratini2025data}. These challenges highlight the urgent need for privacy-preserving approaches to cross-border health data analysis.

International infectious disease surveillance is a critical aspect of global health security, aimed at detecting and responding to outbreaks rapidly \cite{lu2024developing}. Current surveillance systems involve a network of national and international organizations, including the World Health Organization (WHO) \cite{whodata}, the European Centre for Disease Prevention and Control (ECDC), the Chinese Center for Disease Control and Prevention (China CDC) \cite{chinacdcdata}, and national entities like the Portuguese Directorate-General of Health (DGS). These organizations use a combination of traditional epidemiological methods and modern technologies, such as real-time data collection, genomic sequencing, and artificial intelligence (AI), to monitor disease trends and potential outbreaks.

However, global disease surveillance faces numerous challenges. Firstly, data sharing between countries can be hindered by political and privacy concerns, leading to delays in outbreak detection and response \cite{tacconelli2022challenges,edelstein2018strengthening,lencucha2021trust}. Secondly, disparities in healthcare infrastructure and resources result in inconsistent surveillance capabilities in different regions. Low-income countries often lack the necessary resources and trained personnel for effective monitoring \cite{kaushik2025challenges}. Additionally, emerging diseases and pathogens pose a constant threat, requiring continual adaptation of surveillance systems \cite{mcarthur2019emerging,zhou2025integrated}. Lastly, misinformation and public mistrust can undermine efforts to control the spread of infectious diseases \cite{van2022misinformation,chowdhury2023understanding}. Addressing these issues is crucial for improving global health security and effective disease control\cite{czerniak2023scoping}.

For example, during the early stages of the COVID-19 pandemic, several countries hesitated to share granular epidemiological data due to concerns over privacy, national security, and political control \cite{kalkman2022patients}. Similarly, differing interpretations of the EU General Data Protection Regulation (GDPR) have complicated transnational research efforts, even within Europe. These situations underscore the tension between public health urgency and privacy protection \cite{duarte2024stochastic}.

In recent years, big data-driven AI technologies have shown great potential in fields such as finance, healthcare, urban planning, and autonomous driving. However, privacy protection issues often become the main obstacle to their widespread application. Regular machine learning (ML) methods require service providers to centrally collect user data containing sensitive information, such as age, race, and disease information, to train models. This data collection process can directly or indirectly expose the privacy of users \cite{freitas2023data}. If this sensitive information is leaked or misused, it can be threatening to users' personal safety, reputation, and financial security.

As user concerns about privacy protection increase, the willingness to share data decreases, while the development of AI technologies relies on large amounts of data. This has intensified the contradiction between data privacy protection and large-scale data utilization. In this context, cross-border collaboration on the use of privacy protection, data analysis, and sharing technologies has become an innovative practice to resolve this contradiction. The use of Federated Learning (FL) approaches provides a new solution to the conflict between privacy protection and data utilization \cite{kairouz2021advances}. FL, an emerging distributed ML paradigm, aims to collaboratively train models across multiple data holders, such as mobile devices, organizations, or data centers, without directly exchanging or centralizing raw data, thereby enabling data-driven intelligence while preserving data privacy and regulatory compliance \cite{AGRAWAL2022346, alazab2021federated}.


By keeping raw data local and exchanging model parameters instead, FL can substantially reduce the need for centralized data sharing while still enabling collaborative model training. However, FL does not entirely eliminate privacy risks, as model updates may still leak sensitive information through inference attacks or other forms of information disclosure. In the context of epidemic monitoring, FL therefore offers a promising framework for privacy-preserving cross-border collaboration, but its deployment should still be accompanied by careful consideration of residual privacy and security risks.

This study is motivated by the need to explore privacy-preserving, collaborative frameworks for international epidemiological modeling. FL has emerged as a viable solution to train machine learning models across decentralized data sources without compromising data privacy. However, its real-world applicability across heterogeneous jurisdictions remains underexplored. By deploying a cross-border FL experiment across Portugal, Macau Special Administrative Region (SAR), and Guangdong, we assess the feasibility of FL under the real constraints of global disease surveillance.

The FL project between Guangdong, the Macau SAR, and Portugal is an innovative action implemented under this technological background to meet epidemic monitoring needs. Through this project, the three regions achieved collaborative model training by exchanging encoded intermediate parameters, greatly reducing the risk of data leakage.


\subsection*{Legal and Ethical Considerations in Cross-Border Collaboration}

While this study demonstrates the technical feasibility and predictive effectiveness of cross-border FL, it also operates within the complex landscape of data sovereignty, legal compliance, and ethical regulation. Each participating region, Portugal, Macau SAR, and Guangdong, has its own legal frameworks governing personal health data usage, storage, and transmission.

To navigate these challenges, we intentionally designed the FL architecture to comply with data localization and privacy-preserving principles, ensuring that raw data remains within each jurisdiction at all times. Only the model parameters (not raw or identifiable data) are exchanged during training, thus reducing the risk of violating data protection laws such as the EU GDPR in Portugal and corresponding national regulations in China and Macau SAR.
Prior to the start of the project, it was guaranteed by the data-owning entities in the three regions that the datasets were deposited in public repositories (WHO, ECDC, and China CDC), allowing the centralized model to be tested during the experiments as a reference. This experience demonstrates that, in realistic scenarios where sensitive data cannot be shared across institutions or borders, FL enables the development of reliable models and predictions without compromising data privacy. It thus ensures compliance with data protection regulations in all participating countries, while respecting data sovereignty and ethical collaboration principles.

By taking these precautions, this work addresses the core legal and ethical concerns that often hinder international AI collaboration in healthcare, offering a replicable framework that balances scientific collaboration with strict adherence to data governance requirements.

This design also ensured that no cross-border transfer of raw data occurred, aligning with applicable regulatory frameworks including GDPR and relevant Chinese data protection policies.

\vspace{5pt}
\section*{Related work}

Recent studies have demonstrated that FL not only enhances model performance but also fosters collaboration across institutions, leading to more timely and accurate insights into pandemic trends.

FL enables the integration of heterogeneous datasets, capturing varying demographic, geographic, and temporal factors that influence the spread and impact of the virus. It has been successfully applied to a range of tasks, including COVID-19 diagnosis using chest X-ray images \cite{darzi2024comparative, giuseppi2022decentralised, chest_xray,chest_xray2,xray1arlindo,xray2arlindo}, assessing public health risks \cite{dayan2021federated, vaid2021federated}, and optimizing the allocation of medical resources \cite{nguyen2022federated, hamood2023advancing, samuel2022iomt}.  Additionally, it facilitated cross-hospital learning from electronic health records (EHRs), allowing hospitals to collaboratively enhance diagnostic and treatment protocols while keeping sensitive data secure and decentralized.

Besides the justifiable application to medical images data, due to its inherent privacy concerns, FL has also been applied to time-series data related to COVID-19 and other pandemics, although to a lesser extent.
Table~\ref{tab:covid_and_fl} characterizes some notable examples, which include the construction of COVID-19 vulnerability maps using data from self-reporting apps in the United States \cite{chen2021privacy}, the prediction of people crowd movement during epidemic outbreaks in China \cite{wang2022privacy}, and the estimation of COVID-19 cases in different provinces in India \cite{pandianchery2023centralized}.
However, to the best of our knowledge, no other work has dealt with inter-country COVID-19 related time-series data.

As the field continues to evolve, FL is poised to play a crucial role in the development of intelligent epidemic forecasting systems and in the ongoing global effort to manage and mitigate the impact of COVID-19 and other infections.

\begin{figure*}[htb]
    \centering
    \includegraphics[width=0.8\textwidth]{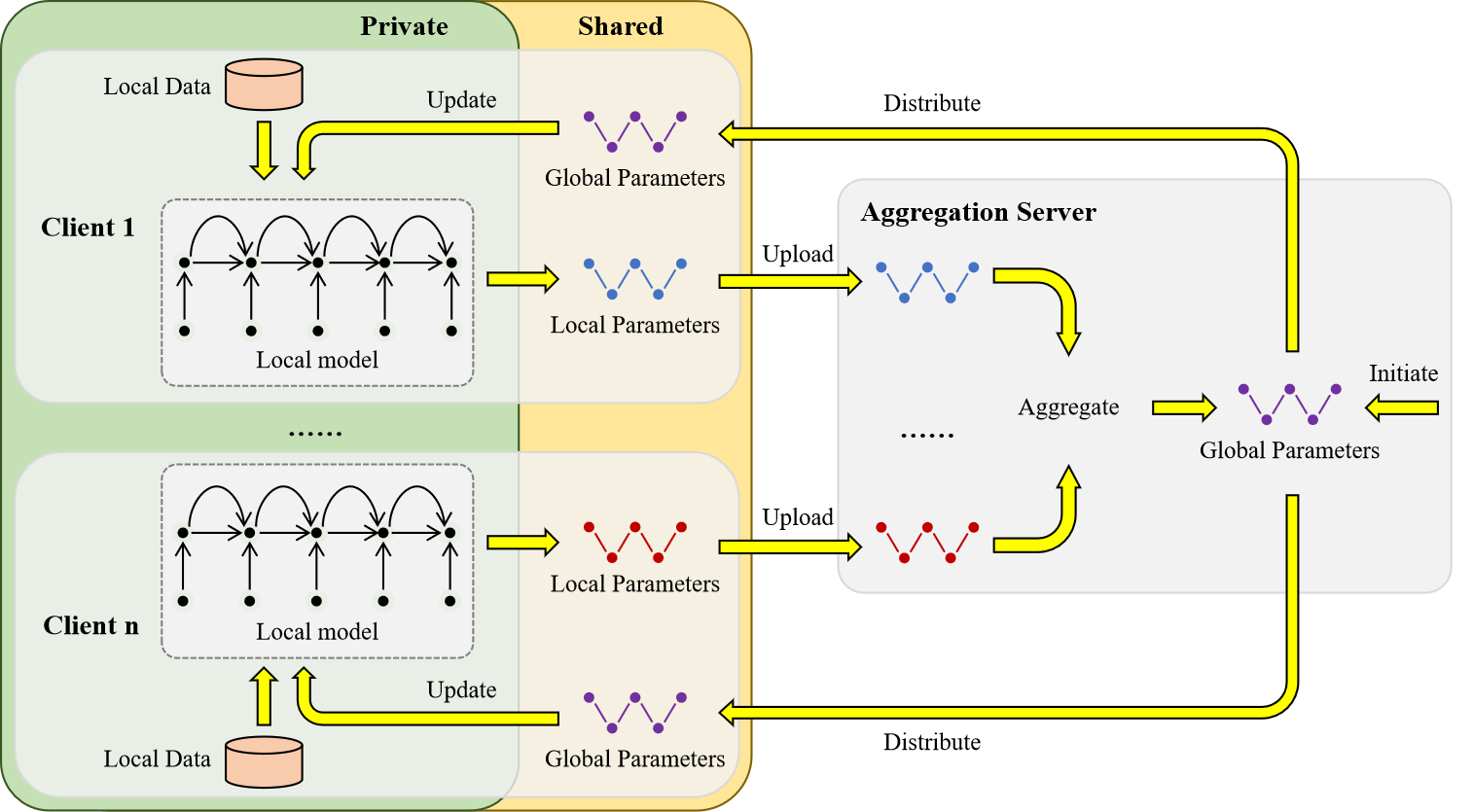}
    \caption{Workflow for a generic server-based FL training process: (1) global model initialization, (2) distrubute global parameters to clients, (3) model trained on local data, (4) clients upload parameters to aggregation server, (5) aggregation of local parameters into new global parameters, (6) repeat steps 2-5.}
     \label{fig:fl_workflow}
\end{figure*}

Several FL frameworks and tools have been developed in recent years, with some popular implementations including TensorFlow Federated \cite{tff}, Flower \cite{flower}, FedML \cite{fedml}, FATE \cite{fate}, AutoFL \cite{autofl}, FLRA \cite{flra}, PySyft \cite{pysyft}, Nvidia FLARE \cite{flare}, OpenFL \cite{openfl} and FLUTE \cite{flute}.
Most of these implementations are based on a server-based approach consisting of a main server interested in creating the ML model and several clients which own their private unmovable data.
The training process is then carried out by the following workflow, also depicted in Fig.~\ref{fig:fl_workflow}: (1) the server initializes a global model or fetches it from a random client; (2) a copy of this global model is sent to every client; (3) each client trains this global model on its own data producing a local model; (4) each client sends back its local model to the server; (5) the server merges together all these local models using an aggregation procedure, producing a new global model; (6) steps 2 to 5 are repeated for a specified number of rounds until convergence.

While prior studies have explored FL for COVID-19-related time-series prediction, most are limited to single-country settings or simulated environments with homogeneous infrastructure. In contrast, our work evaluates a real-world cross-border deployment spanning Portugal, Macau SAR, and Guangdong, explicitly addressing heterogeneity in public health interventions, data distributions, and computational resources. Furthermore, our experimental setup incorporates geographically distributed nodes with differing hardware configurations, reflecting realistic operational constraints. This distinguishes our study as, to the best of our knowledge, one of the first studies evaluating federated time-series forecasting across multiple countries under real-world deployment constraints.



\begin{table*}[!ht]
\small\sf\centering
\caption{\label{tab:covid_and_fl}Examples of federated learning applications for COVID-related time-series prediction. These studies used different models but more or less the same metrics for the models evaluations. Models: Long short-term memory (LSTM); Convolutional Neural Networks and LSTM (CNN-LSTM); CNN-based model. Metrics: Root Mean Square Error (RMSE), Mean Absolute Error (MAE) and Mean Absolute Percentage Error (MAPE).}
\begin{tabular}{clllll}
    \toprule 
    Ref & Year & Prediction target & Datasets & Models & Metrics \\
    \midrule 
     \cite{chen2021privacy} & 2021 & Vulnerability risk map & 88 neighborhoods of Houston & LSTM & RMSE \\
     \cite{wang2022privacy} & 2022 & Crowd flow & Crowd travel trajectories & CNN-LSTM & RMSE, MAE, MAPE \\
     \cite{pandianchery2023centralized} & 2023 & Active cases per day & 36 provinces in India & CNN-based model& RMSE, MAE \\
    \bottomrule 
\end{tabular}
\end{table*}


\section*{Materials and methods}

This study was conducted between March and November 2024 as a multi-center experimental study at three sites: INESC-ID (Lisbon, Portugal), the Macau University of Science and Technology (Macau SAR, China), and Guangzhou Medical University (Guangzhou, China). The methodology and reporting were guided by the Transparent Reporting of a Multivariable Prediction Model for Individual Prognosis Or Diagnosis-Artificial Intelligence (TRIPOD-AI) statement \cite{tripod+ai}.

\subsection*{COVID-19 dataset and preprocessing}

\begin{figure*}[!ht]
    \centering
    \includegraphics[width=1.0\textwidth]{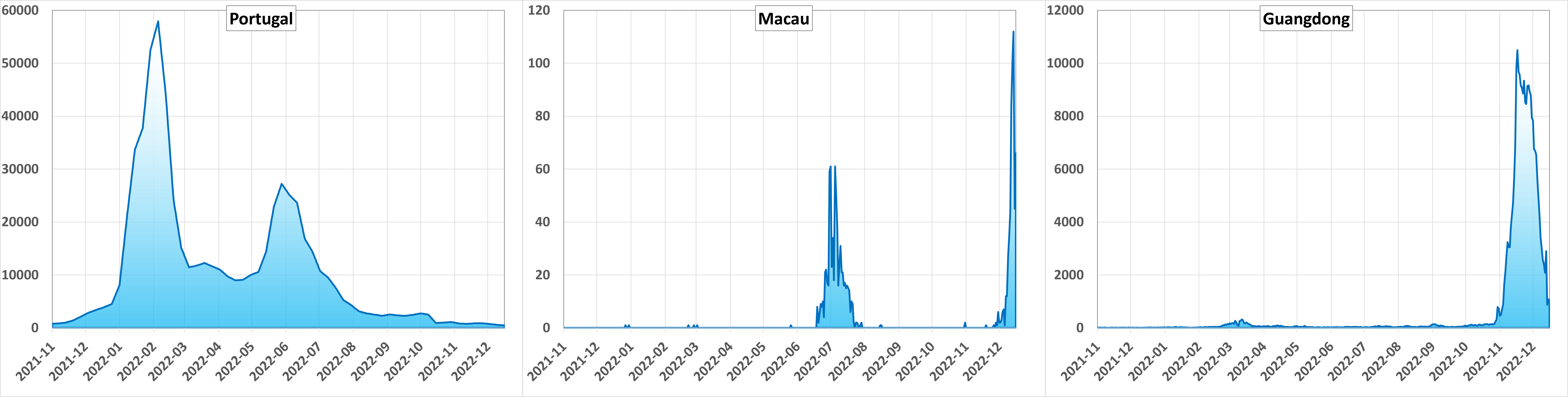}
    \caption{Raw data corresponding to the reported new COVID-19 cases for the three locations (from left to right: Portugal, Macau SAR, Guangdong) in the time period ranging from 2021-11-24 to 2023-01-08, a total of 14 months.}
    \label{fig:input_data}
\end{figure*}

We explore the COVID-19 pandemic data to build a model for effective prediction of infectious diseases over time. By resorting to historical epidemic data from various regions, we aim to enhance the accuracy of the predictive models explaining the spread of the virus across different temporal windows. For this purpose, data from two different countries, Portugal and China, and more specifically from three geographical regions, Portugal (all over the country), Macau SAR, and Guangdong, were collected. Two distinct data sources were used, the WHO \cite{whodata} and the China CDC dataset \cite{chinacdcdata}. This data reflects the daily number of reported cases ranging from 2021-11-24 to 2023-01-08 for each one of the regions, whose profiles are presented in Fig.~\ref{fig:input_data}, summing up to a total of 1233 samples. This initial date corresponds to the first reported Omicron variant case \cite{omicron}.

In each experiment, close to 80\% of the samples from each region were reserved for training, while the remaining samples were used for testing.
Training data covers the period ranging from 2021-11-24 to 2022-10-17, the first 11 months, 328 days.
The test split was not based on randomly shuffled data, but taking the chronologically more recent samples, from 2022-10-18 onwards (83 days), corresponding to the time period of the most severe COVID-19 outbreak in China in late 2022.
This enabled us to assess our models regarding the prediction of new continued epidemic events and beyond the basic interpolation between time steps.

\begin{figure*}[ht]
    \centering
    \includegraphics[width=1.0\textwidth]{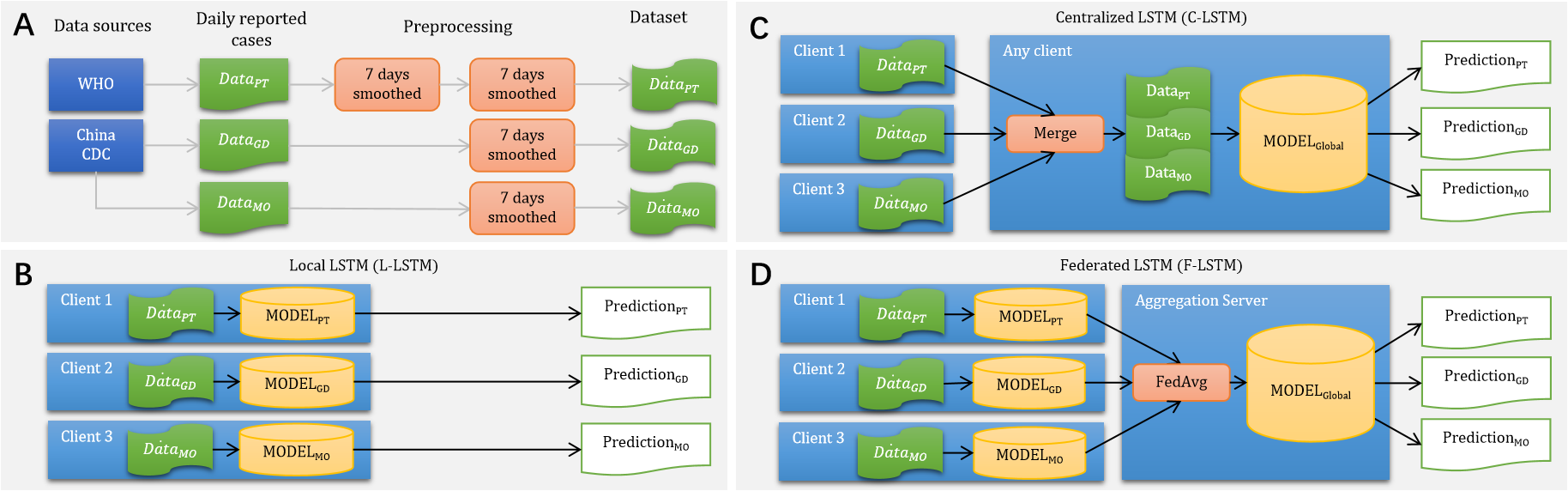}
    \caption{Experimental design of the project, including \textbf{(A)} data processing and the three model designs: \textbf{(B)} Local, \textbf{(C)} Centralized and \textbf{(D)} Federated.}
    \label{fig:architecture}
\end{figure*}

To reduce errors caused by delays in the detection and reporting of COVID-19 cases, we applied a 7-day moving average to the raw data. 
In case of Portugal, where the new confirmed cases were only reported weekly, we applied the 7-day moving average twice to make the data appear as daily increments. Fig.~\ref{fig:architecture}-A illustrates this preprocessing step. Additionally, due to the different geopolitical nature of the three regions, comprising a county-like region, a district and a whole country, the data for each region was scaled to the unitary range. This is required when assembling data from all the three regions together in the same model, as is the case of the FL approach, since they feature uneven peak values ranging from tens to tens of thousands.

\subsection*{Federated Learning topology and design}

The FL architecture typically comprises two core components: a central aggregation server and several distributed clients, also referred to as participants \cite{flopportunitieschallenges, FederatedMachineLearning}.


The tripartite computational infrastructure incorporates multiple geographically distributed hubs in Guangzhou, Macau SAR, and Lisbon, each serving as a client for data processing and model training, coordinated by an additional node placed in Lisbon working as the central aggregation server. By adopting a FL methodology, these centers collaborate to jointly optimize a shared predictive model while maintaining data residency within their respective jurisdictions. The architecture consists of interconnected yet independent computing environments, wherein each center trains a local model on its confined datasets derived from regional infectious disease surveillance activities. Subsequently, they communicate only model updates or gradients rather than raw data, thus respecting data localization laws and ethical considerations.


Our FL implementation (Fig.~\ref{fig:architecture}-D) employs the highly adaptable Flower framework \cite{flower} for the secure sharing of model weights. The framework's modularity, scalability, inherent security features, extensive monitoring capabilities, and support for customization facilitate efficient, secure, and contextually tailored model training. As aggregation method, the federated averaging (FedAvg) strategy \cite{fedavg} was chosen, which is based on a weighted average of the parameters dependent on the number of samples on each client.
At the start of training (Round 0), the global model was initialized on the central server using parameters sampled from one participating client and subsequently broadcast to all nodes to ensure a consistent initialization across sites.

\subsection*{Model architecture and design}

Long Short-Term Memory (LSTM) networks \cite{lstm} are a special type of Recurrent Neural Networks (RNNs) designed to address the vanishing gradient problem encountered in traditional RNNs. They introduce memory cells with gates (forget, input, and output) that regulate the flow of information, enabling LSTMs to capture long-term dependencies effectively. This capability makes LSTMs particularly suitable for tasks involving sequences, such as language modeling, time series prediction, and natural language processing, where understanding the context from the past is crucial.

Although more advanced architectures for time-series forecasting, such as Transformers \cite{transformers}, have recently shown strong performance, we selected LSTMs for this study due to their simpler architecture, smaller parameter footprint, and lower computational and communication demands in the federated setting. These properties are particularly relevant in our cross-border deployment, where the participating nodes had heterogeneous hardware resources, and repeated model exchange was required across training rounds. Furthermore, given the relatively limited size of the dataset and the short temporal input window used in this study, the LSTMs provided an appropriate balance between representational capacity, training stability, and deployment efficiency. A systematic empirical comparison with transformer-based architectures remains an important direction for future work.

To better understand the pros and cons of FL in comparison to traditional ML method, we experimented with three distinct approaches: a localized LSTM (L-LSTM), which produces three different models, one for each country, where each model is only trained with data from its own country in isolation (Fig.~\ref{fig:architecture}-B); a centralized LSTM (C-LSTM), where a single model is trained with data from all the countries collectively, data is merged, corresponding to each country having local access to data from every other country (Fig.~\ref{fig:architecture}-C); a federated LSTM (F-LSTM), where the global model is trained on three different servers in different geographies and aggregated on an independent external server using the FL approach, as detailed in Fig.~\ref{fig:fl_workflow}, without any data leaving its corresponding geography (Fig.~\ref{fig:architecture}-D). For this particular study, it was possible to test the centralized model because the three regions had already agreed to publicly deposit the data with the WHO, the ECDC, and the China CDC. 

The three local L-LSTM models were used as a baseline. The centralized model, C-LSTM, was used as the gold standard, reflecting the fact that in many situations this would be the best possible model but it cannot be created because data cannot be put together in a centralized location, due to privacy issues and other restrictions. The FL experiment was deployed across four servers, two in Lisbon, one in Guangzhou and one in Macau SAR, explicitly testing real conditions for cross-border federated learning. In our case, the FedAvg method is adopted to combine all of the model updates received from the clients.  FedAvg takes a weighted average of model updates, weighted by the number of examples each client used for training. All initial raw data were subjected to the same pre-processing steps (Fig.~\ref{fig:architecture}-A).

\subsection*{Hardware configuration}

The hardware configuration of the FL experimental setup is detailed in Table~\ref{tab:hardware_config}. The distributed system consists of three clients located in Lisbon, Portugal (hosted by INESC-ID), Guangzhou, China (hosted by GMU), and Macau SAR, China (hosted by MUST), with an aggregation server co-located with Client No.1 at INESC-ID in Lisbon. Each node was equipped with varying computational resources to simulate realistic heterogeneous edge computing environments: Client No.1 and the aggregation server were allocated 2 virtual CPU (vCPU) cores (Intel Xeon Silver 4214 @ 2.2GHz) with 2GB RAM, while Client No.2 featured higher specifications (4 cores, Intel Xeon Platinum 8269CY @ 2.5GHz, 8GB RAM) representing a more capable edge device. Client No.3 maintained intermediate resources (2 cores, Intel Xeon E5-2609 @ 1.9GHz, 8GB RAM). All nodes ran Ubuntu 22.04 LTS (except Client No.1 using the non-LTS version). Notably, Client No.3 at MUST utilizes VMware ESXi virtualization, while INESC-ID and GMU employ Kernel-based Virtual Machine (KVM). The storage configuration ranged from 25GB for lightweight nodes to 500GB for data-intensive processing at Client No.3.

\begin{table*}[htbp]
\small\sf\centering
\caption{\label{tab:hardware_config}Hardware configuration of the FL platform}
\begin{tabular}{lcccc}
\toprule
\textbf{Specification} &
\textbf{\textbf{Client No.1}} & \textbf{Client No.2} & \textbf{Client No.3} & \textbf{Aggregation Server}\\
\midrule
\textbf{Location} & Lisbon, Portugal & Guangzhou, China & Macau SAR, China & Lisbon, Portugal \\
\textbf{Institution} & INESC-ID & GMU & MUST & INESC-ID \\
\textbf{Memory} & 2GB & 8GB & 8GB & 2GB \\
\textbf{Storage} & 25GB & 40GB & 500GB & 25GB \\
\textbf{OS} & Ubuntu 22.04 & Ubuntu 22.04 LTS & Ubuntu 22.04 LTS & Ubuntu 22.04 LTS \\
\textbf{Virtualization} & KVM & KVM & VMware ESXi & KVM \\
\textbf{CPU} & Intel Xeon Silver 4214 & Intel Xeon Platinum 8269CY & Intel Xeon E5-2609 & Intel Xeon Silver 4214 \\
\textbf{Allocated vCPU} & 
2 Cores @ 2.2GHz & 4 Cores @ 2.5GHz & 2 Cores @ 1.9GHz & 2 Cores @ 2.2GHz \\
\bottomrule
\end{tabular}
\end{table*}

While the hardware infrastructure was heterogeneous, all sites used the same version of the Flower FL framework to ensure consistency in the orchestration of training rounds and aggregation. Despite this standardized software layer, the underlying datasets were heterogeneous in structure and size, reflecting the local epidemiological and reporting practices of each region. To account for this variability, we harmonized the data preprocessing by applying a 7-day rolling average window, allowing for comparability in model inputs across clients. This configuration shows that the federated setup can operate under real-world constraints, where differences in hardware and dataset structure are the norm.

\subsection*{Model implementation}

In this study, we employed a series of specialized software packages to support the FL framework and address data processing needs.
The implementation was developed in Python 3.12, with the PyTorch library \cite{pytorch} being used to build all the LSTM models. The Flower framework \cite{flower} version 1.7.0 was used to train and process the F-LSTM model.

All models were designed to learn the COVID-19 temporal evolution based on a one week window, i.e. each model takes as input the number of new daily cases from days 1 to 7 and predicts the number of cases of the 8th day.
After this, the model architecture was updated to extend the inference to the range of 1 to 7 days ahead, in order to assess the limits of the model's predicting capabilities.
In more detail, the LSTM model implementation uses an input size of 7, representing the daily new confirmed COVID-19 cases from the past 7 days as features. The hidden state size was set to 64, where each LSTM layer captures temporal dependencies using 64 neurons. The model consists of 2 stacked LSTM layers for learning complex sequence patterns. A fully connected layer maps the final hidden state to an output size of 1, predicting the next day's number of new cases. Since daily infection counts cannot be negative, we applied a non-negativity constraint $F_t = \max(0, F_t)$ to ensure the physical plausibility of the results, where $F_t$ denotes the forecasted value at time $t$. The LSTM architecture, comprising two layers with a hidden size of 64, was selected based on preliminary experiments balancing predictive performance and model complexity. Deeper or larger models did not yield consistent improvements while increasing the risk of overfitting and communication overhead in the federated setting. This configuration provided stable training and good generalization across regions.

L-LSTM and C-LSTM models were trained for 1000 epochs using an early-stopping strategy with a patience factor of 10 epochs in order to prevent over-fitting.
F-LSTM model was trained for 200 rounds using one epoch per round in each client, and taking the top scoring round as detailed in section titled "Number of training rounds".

\subsection*{Statistical Analysis}

To evaluate and compare the predictive performance of the different models, we relied on three popular metrics, Root Mean Square Error (RMSE), Mean Absolute Error (MAE) and Symmetric Mean Absolute Percentage Error (SMAPE). The RMSE provides a measure of the magnitude of the error. It is particularly useful when large errors are particularly undesirable. The MAE is a straightforward measure of the accuracy of a model, representing the average magnitude of errors in a set of predictions, without considering their direction. The SMAPE is commonly used in time series forecasting and regression analysis to measure the accuracy of predictions based on percentage errors. SMAPE is a normalized version of the Mean Absolute Percentage Error (MAPE), making it symmetric by considering the absolute values of both the actual and forecasted values in the denominator. The formula for RMSE, MAE and SMAPE are given by:

\begin{equation}
RMSE = \sqrt{ \frac{1}{n}  \sum_{t=1}^{n} (F_t - A_t)^2}
\end{equation}

\begin{equation}
MAE = \frac{1}{n} \sum_{t=1}^{n} |F_t - A_t|
\end{equation}

\begin{equation}
SMAPE = \frac{100}{n} \sum_{t=1}^{n} \frac{|F_t - A_t|}{\left( |A_t| + |F_t| \right)}
\end{equation}

In these formulas, \( n \) represents the number of forecasted points, \( F_t \) denotes the forecasted value at time \( t \). \( A_t \) denotes the actual value at time \( t \), and \( | \cdot | \) denotes the absolute value.

\begin{equation}
nRMSE_{s} = \frac{RMSE_{s}}{\max(\{x_{s,t} \mid x_{s,t} \in X_s\})}
\end{equation}

\begin{equation}
nMAE_{s} = \frac{MAE_{s}}{\max(\{x_{s,t} \mid x_{s,t} \in X_s\})}
\end{equation}

Since Guangdong, Macau, and Portugal all have different population numbers and public health measures, normalization was performed so that the RMSE and MAE can be compared across different regions. In the above formulas, $X_s$  denotes the total set of daily reported new confirmed cases for region $s$, among Macau, Guangdong or Portugal, with \( x_{s, t} \) corresponding to the specific count at date \( t \).

\subsection*{Network communication metrics}

Since FL depends on model sharing between distributed remote nodes, the network communication consumption and overheads, including the size of data transferred and the training time, was also considered in the evaluation. The Flower framework implementation uses TCP as the transport protocol, and each TCP connection consumes network resources during initialization and maintenance, including the three-way handshake, transmission acknowledgment packets, and handling possible connection resets. Additionally, FL data packets support encryption and integrity verification, which increases the data volume and processing time for each transmission. To this end we used the Wireshark software \cite{wireshark} to capture the network log during the training process, to later analyze it regarding these network metrics.

\section*{Results}

\subsection*{Prediction accuracy of daily new cases}

\begin{figure*}[!ht]
    \centering
    \caption{Comparative 7-day ahead prediction results from the L-LSTM, C-LSTM, and F-LSTM models. The red line represents the observed real data; the green and blue lines represent the L-LSTM and C-LSTM predictions, respectively; the purple line indicates the point predictions from the proposed F-LSTM model; and the shaded area denotes the 95\% prediction intervals (PIs) for the F-LSTM model, which are derived from the empirical distribution of test-set residuals with a non-negativity constraint to quantify forecast uncertainty.}
    \begin{subfigure}{0.49\textwidth}
        \includegraphics[width=\textwidth]{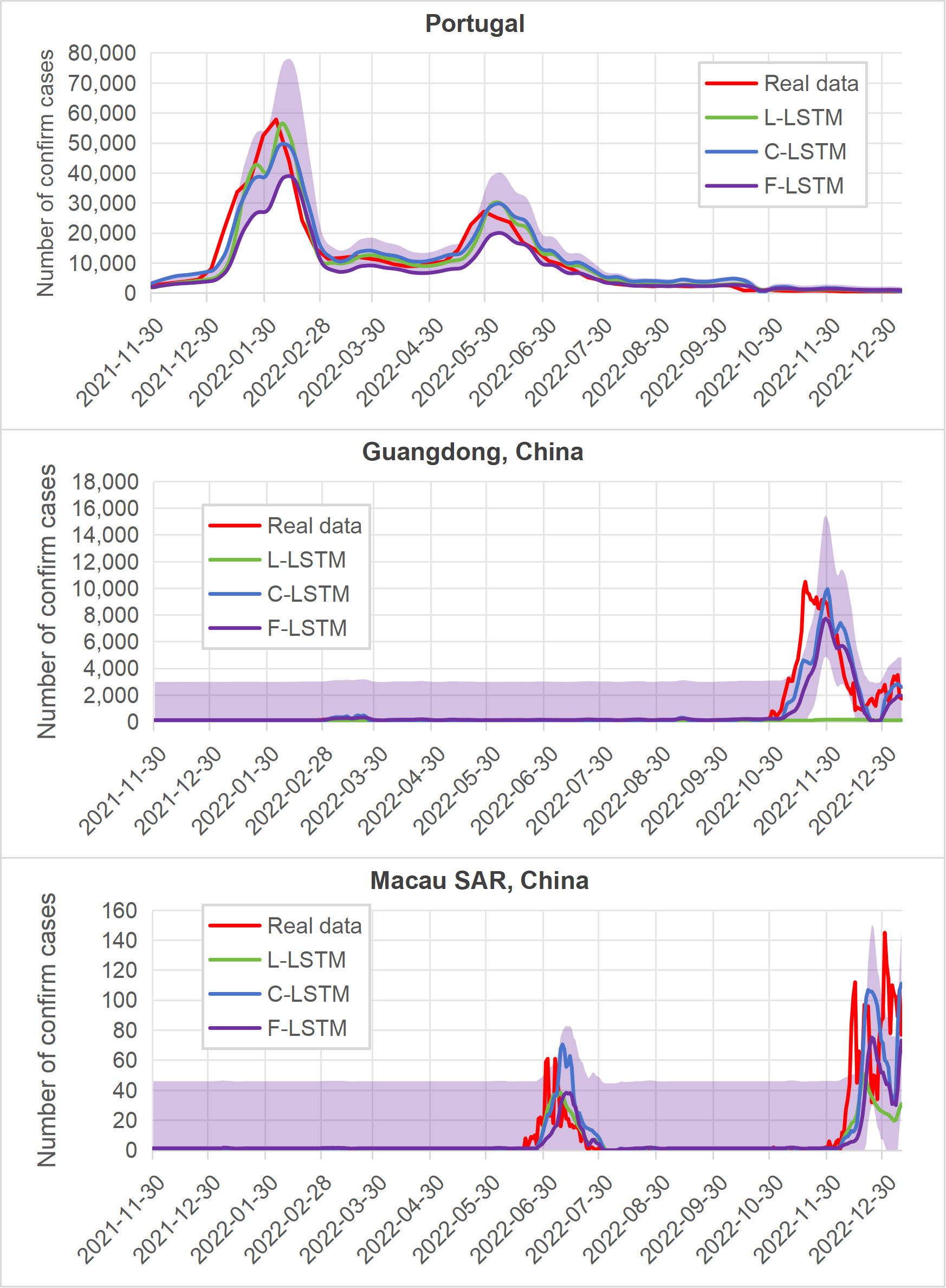}
        \caption{Results for the 7-day ahead prediction across the three regions for the entire dataset range (Nov 2021 – Jan 2023).}
        \label{fig:3vscomparison}
    \end{subfigure}
    \begin{subfigure}{0.49\textwidth}
        \includegraphics[width=\textwidth]{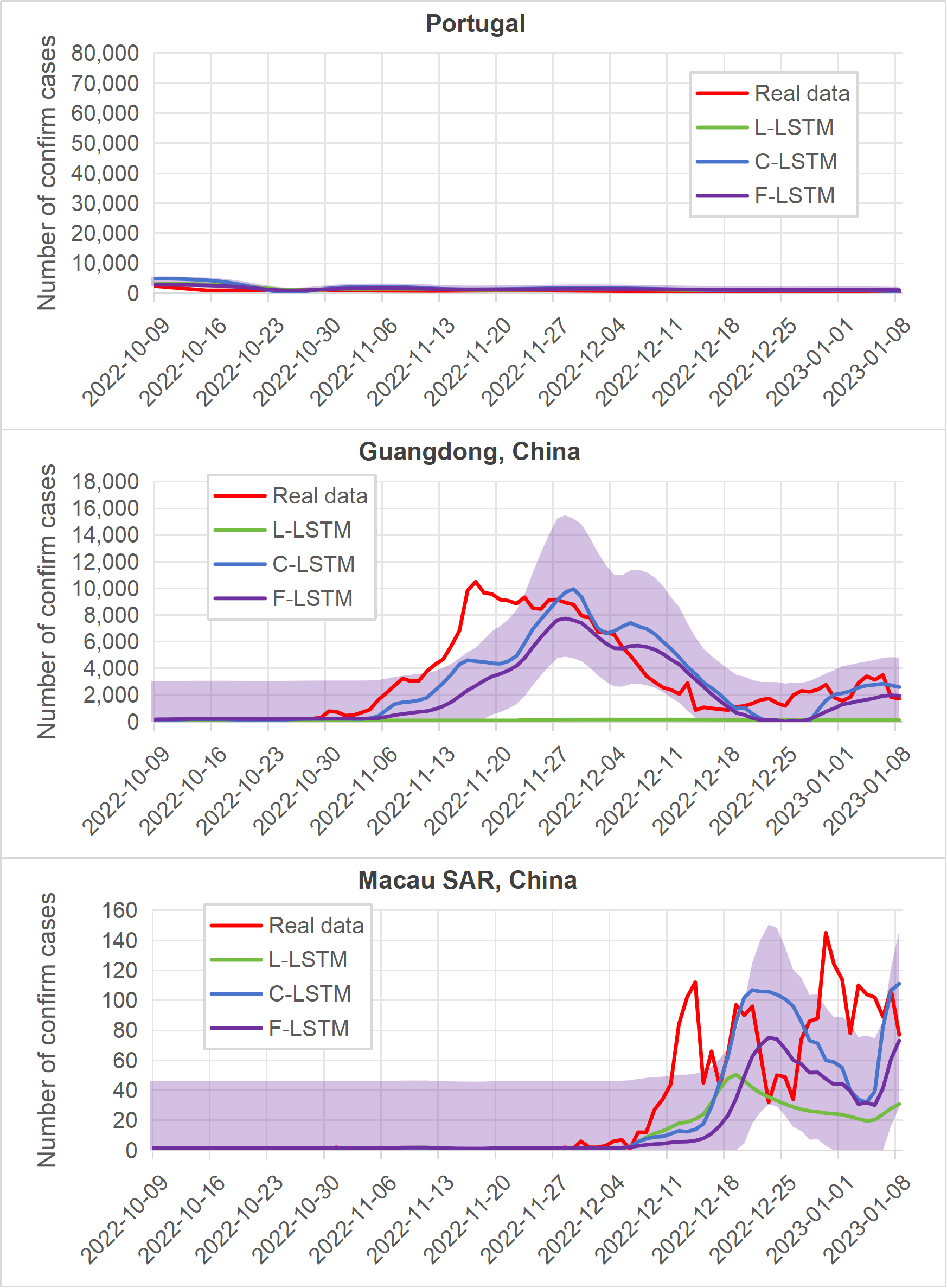}
        \caption{Zoomed-in view of the 7-day ahead prediction over the last three months of data (Oct 2022 – Jan 2023).}
        \label{fig:7daysprediction}
    \end{subfigure}
    \label{fig:predictions}
\end{figure*}

The daily new COVID-19 infection numbers across Portugal, Guangdong and Macau SAR predicted by the three models for the testing interval corresponding to the final three months of data are shown in Fig.~\ref{fig:predictions}. These plots display the period relative to the test partition only, corresponding to the last months of the full dataset, not previously seen in the training process.

Although the number of cases in Portugal remained steadily low during this period, the situation in China differed greatly, presenting an unprecedented high number of cases in both Macau and Guangdong.
Both the C-LSTM and F-LSTM models were able to match these unseen deviations more accurately than the local models, as confirmed by the blue and purple lines matching the dashed red line in Fig.~\ref{fig:7daysprediction}, while L-LSTM performed inadequately in all cases, with predictions far-off from the real values, as seen by the distance between the green and red dashed lines in the same figure.
This under-performing model overestimated the numbers in Portugal while underestimating them in China. Nevertheless, it provided better predictions for Macau as it had already observed a moderate rise in cases before, in opposite to Guangdong where the training data, as well as the test predictions, were mostly flat.
The C-LSTM and F-LSTM models were much more competent in China as they were previously exposed to the data from Portugal and had experienced peaks featuring a high number of cases before.

A more detailed look at the metrics for this evaluation period, presented in Table~\ref{tab:performance}, also confirms these observations, with a clear advantage of the last two models over L-LSTM across all scores.
F-LSTM provided comparable results to C-LSTM, with the former being slightly more efficient than the later, except in the Guangdong region, possibly due to the majority of the training samples being very close to zero.

In particular, the federated model consistently achieves markedly lower prediction errors than the local models, reflecting the benefit of cross-border knowledge integration. When averaged across regions, the federated approach reduces nRMSE from 0.240 to 0.028, corresponding to an 88.3\% relative improvement over the local baseline, while maintaining performance comparable to, and slightly better than, the centralized model. These results demonstrate that federated learning can effectively approximate the performance of centralized training without requiring data sharing.


Regarding the forecast of a more distant time frame into the future, Fig.~\ref{fig:3vscomparison} shows the results for the F-LSTM predictions ranging from 1 day ahead to 7 days ahead.
The task of predicting a full week ahead degrades along the course of the week as expected, but still maintains a reliable accuracy even at the 7th day.
The model seems to progressively underestimate the maximum values in Portugal and overestimate the peaks in both China regions.

\begin{table}[ht]
\small\sf\centering
\footnotesize
\captionsetup{type=table}
\captionof{table}{Performance metrics comparison for 1-day ahead and 7-day ahead predictions across three regions. The best-performing metrics are highlighted in bold (lower values indicate higher accuracy). All metrics are calculated based on normalized data to ensure comparability across regions with different population scales.}
\label{tab:performance}
\begin{tabular}{l|ll|rrr}
    \toprule 
    Ahead days  & Location    & Method & nRMSE & nMAE & SMAPE \\
    \midrule 
    1 & Portugal   & L-LSTM & 0.051          & 0.051          & 66.756 \\
      &            & C-LSTM & 0.010          & 0.010          & 28.133 \\
      &            & F-LSTM & \textbf{0.001} & \textbf{0.001} & \textbf{3.086} \\
    
      & Guangdong  & L-LSTM & 0.486          & 0.359          & 100.000 \\
      &            & C-LSTM & \textbf{0.017}          & \textbf{0.012}          & \textbf{3.083} \\
      &            & F-LSTM & 0.030 & 0.022 & 6.034 \\
      & Macau      & L-LSTM & 0.182          & 0.095          & \textbf{18.385} \\
      &            & C-LSTM & 0.064          & 0.036          & 28.087 \\
      &            & F-LSTM & \textbf{0.053} & \textbf{0.026} & 39.748 \\
      & Average    & L-LSTM & 0.240          & 0.168          & 61.714 \\
      &            & C-LSTM & 0.030          & 0.019          & 19.768 \\
      &            & F-LSTM & \textbf{0.028} & \textbf{0.016} & \textbf{16.289} \\
\hline
    7 & Portugal   & L-LSTM & 0.008          & 0.008          & 23.825 \\
      &            & C-LSTM & 0.013          & 0.012          & 31.672 \\
      &            & F-LSTM & \textbf{0.006} & \textbf{0.004} & \textbf{18.914} \\
      & Guangdong  & L-LSTM & 0.481          & 0.354          & 79.425 \\
      &            & C-LSTM & \textbf{0.145}          & \textbf{0.094}          & \textbf{20.268} \\
      &            & F-LSTM & 0.182 & 0.138  & 42.528 \\
      & Macau      & L-LSTM & 0.274          & 0.149          & 48.290 \\
      &            & C-LSTM & \textbf{0.202}          & \textbf{0.108}          & \textbf{43.495} \\
      &            & F-LSTM & 0.244          & 0.128  & 60.678 \\
      & Average    & L-LSTM & 0.254          & 0.170          & 50.513 \\
      &            & C-LSTM & \textbf{0.120}          & \textbf{0.071}          & \textbf{31.812} \\
      &            & F-LSTM & 0.144 & 0.090 & 40.707 \\
    \bottomrule 
\end{tabular}
\end{table}



\subsection*{Number of training rounds}
\label{sec:trainingrounds}

\begin{figure}[!ht]
\centering
\begin{subfigure}{0.5\textwidth}
    \includegraphics[width=\textwidth]{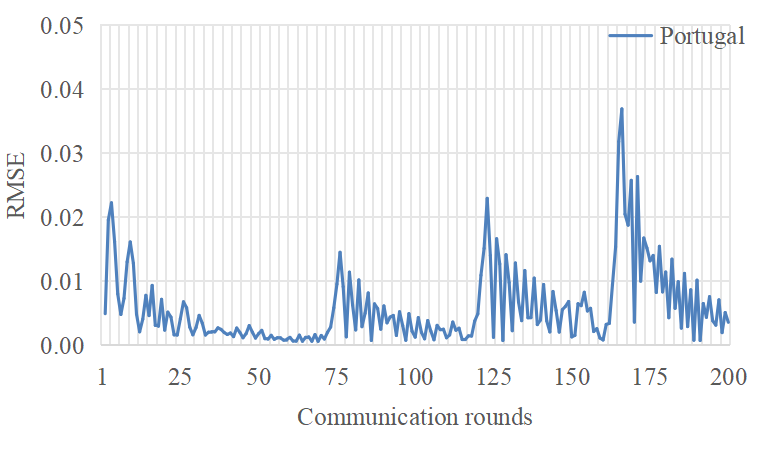}
    \label{fig:figure3a}
\end{subfigure}
\begin{subfigure}{0.5\textwidth}
    \includegraphics[width=\textwidth]{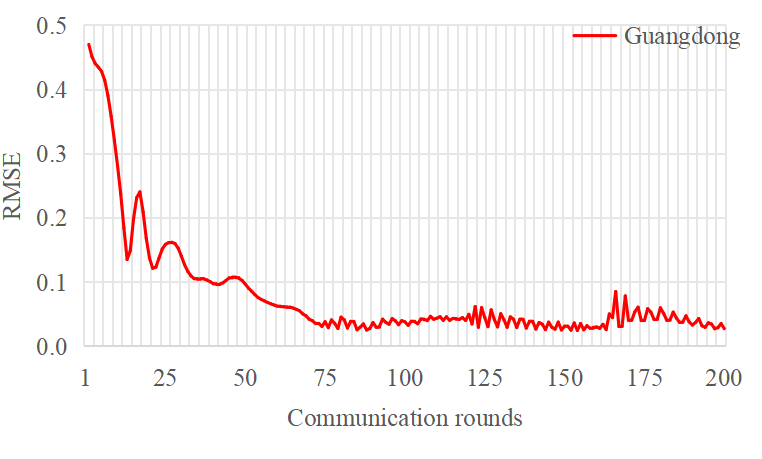}
    \label{fig:figure3b}
\end{subfigure}
\begin{subfigure}{0.5\textwidth}
    \includegraphics[width=\textwidth]{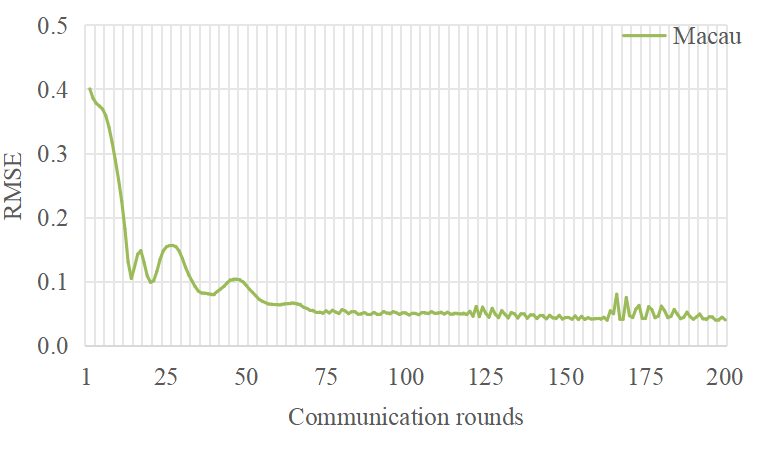}
    \label{fig:figure3c}
\end{subfigure}
\caption{Training loss across 200 rounds for each one of the three locations (from top to bottom: Portugal, Guangdong, Macau). After round 71 the loss starts to oscillate heavily.}
\label{fig:lossvsrounds}
\end{figure}

By experimenting with different numbers of rounds during training, we observed an unexpected behaviour, seen in Fig.~\ref{fig:lossvsrounds}, where the training loss started to oscillate heavily after a certain number of rounds instead of stabilizing over time as in regular localized training.
This irregular phenomenon specific to FL has been previously studied \cite{kundroo2023federated} and is related to the fact the model might start to oscillate between the most important client's local optima, hindering the server to reach the global optimum.
For our results we selected the model corresponding to the number of rounds which minimized the loss function, matching the round before the beginning of the instability. 
We note that selecting the round with the minimum loss is a form of model selection that may bias the reported federated results upward; we did not perform sensitivity analyses over alternative round-selection criteria, nor did we test alternative aggregation or personalization strategies. These analyses remain important future work.
In the future, other solutions to prevent this problem include resorting to a more elaborate FL strategy besides FedAvg or employing an early-stopping mechanism specific for FL \cite{niu2024flrce}, which monitors rounds instead of epochs.

\subsection*{Network communication}

The network traffic size generated during a run of the FL approach is illustrated in Table~\ref{tab:overhead}, captured over a training session with 200 communication rounds, i.e., the initial global model suffers 200 updates from the local clients.
This network traffic primarily depends on factors such as the number of communication rounds between the aggregation server and clients, the size of the model, and the number of clients involved.
The F-LSTM model's parameters data size that gets transferred around is $\sim200$KB.
The server is responsible for sending the model to each one of the three clients two times per round, one time for training the current global model with the client's data, and another for evaluating that round's final global model with each client's data.
The server uploads $\sim240$MB in total, so $240/3=80$MB are sent per client, matching the data received by each one.
In terms of download, the server receives each local model from each client once per round, so $\sim120/3=40$MB per client, which also matches the total value sent by each client.
In each round, two versions of the global model are received by each client, making it $200$KB$*2*200$=$80$MB, which means no significant communication overheads are involved in client downloads.
For client uploads, these amounts are halved 
since the local model is only sent once per round to the server.
Therefore, we can conclude that the data overheads used by the FL strategy are negligible for this type of model size.

\begin{table}
    \small\sf\centering
    \caption{Network data transfers for the F-LSTM training with 200 rounds across four nodes consisting of one server and three clients. Measurements include the number of packets sent/uploaded (Tx), received/downloaded (Rx) and total data transferred (Total).}
    \label{tab:overhead}
    \begin{tabular}{l|ccc}
        \toprule 
        \textbf{} & \multicolumn{3}{c}{\textbf{Size (MB)}} \\
         & $\uparrow$Sent & $\downarrow$Received & Total\\
        \midrule 
        Portugal              &  40.3 &  80.0 & 120.3\\
        Guangdong           &  40.1 &  80.0 & 120.1\\
        Macau               &  40.1 &  80.0 & 120.1\\
        \hline
        Server              & 240.1 & 120.4 & 360.6\\
        \bottomrule 
    \end{tabular}
\end{table}

\begin{table}
\small\sf\centering
\caption{Training times comparison between L-LSTM, C-LSTM, and F-LSTM models across Portugal, Guangdong, and Macau datasets (time in seconds).}
\begin{tabular}{l|cccc}
\toprule 
\textbf{Location} & \textbf{L-LSTM} & \textbf{C-LSTM} & \multicolumn{2}{c}{\textbf{F-LSTM}} \\
& & & Per round & Total \\
\midrule 
    Portugal & 1.090 & 4.212 & 2.237 & 447.4 \\
    Guangdong & 4.071 & 5.876 & 2.237 & 447.4 \\
    Macau & 2.563 & 4.533 & 2.237 & 447.4 \\
\hline
    Average & 2.574 & 4.871 & 2.237 & 447.4 \\
\bottomrule 
\end{tabular}
\label{tab:timings}
\end{table}

Table~\ref{tab:timings} compares the training times of the three different approaches. L-LSTM and C-LSTM training is performed locally offline while F-LSTM is carried out remotely over the internet, therefore incurring in longer times.
To ensure a more balanced comparison, the F-LSTM training on each local client was set to 1 epoch in order to match the number of rounds. This way both the number of rounds and the number of epochs among the different approaches equal the number of updates that the model is subjected to.
All regions have the same amount of samples for L-LSTM since they all correspond to the same time period. Differences in times correspond both to different machine hardware specifications and intrinsic data properties which determine how fast the training converges.
C-LSTM has the exact same data in all regions since it is the union of all three datasets, therefore it is reasonable to assume that it would take three times longer than the L-LSTM time, although in practice this is not so regular.
F-LSTM training is expected to be much slower due to its remote nature, and influenced by several external factors which cannot be controlled, such as ping latency, network infrastructure speeds, traffic congestion, etc.



\section*{Discussion}

\subsection{Key findings}
We explicitly show that, without the Portugal data, the model would not have been able to predict the actual epidemic patterns in Macau and Guangdong observed in late 2022. This highlights the potential advantages of using cross-border FL approaches over localized ones, especially when dealing with sensitive or private data. Simultaneously, these federated efforts encourage and strengthen scientific collaborations between institutions or even countries.

As shown in Figure ~\ref{fig:input_data}, the Omicron outbreak in Portugal started earlier than in Macau and Guangdong, allowing the FL approach to learn the transmission dynamics that helped Macau and Guangdong more accurately anticipate future epidemic trends.

With the continuous input of new data, the model can adjust and optimize in real-time, better reflecting the latest epidemic development trends. For example, if a sudden surge in cases occurs in a particular area, the model can quickly adjust the relevant parameters to predict the possible development direction and impact range of the outbreak. Furthermore, using advanced data analysis techniques, we can further explore deeper correlations and patterns within the data. These technologies can help us identify key predictive indicators from large datasets, such as seasonal variations of the epidemic, the impact of climate factors on the progression, and the contribution of population mobility to virus transmission.

The comprehensive application of FL, real-time data integration, and deep data analysis has significantly enhanced the prediction and early warning capabilities for infectious diseases in the face of the challenge of unclear epidemic transmission patterns. This approach has greatly strengthened the technical support for global public health security, highlighting the practicality and effectiveness of FL in global epidemic monitoring.

\subsection{Implications}
In the experiments, we observed that the training loss began to oscillate after a certain number of rounds, as shown in Figure \ref{fig:lossvsrounds}. This instability, which is specific to FL, occurs when the model oscillates between the local optima of different clients, preventing the server from reaching the global optimum. In our implementation, this issue may have been further exacerbated by the heterogeneous and apparently unbalanced nature of the three regional datasets. Although each client contributed data from the same calendar period, the epidemic trajectories, case magnitudes, reporting practices, and public health interventions differed substantially across Portugal, Guangdong, and Macau SAR. Under such non-IID conditions, the FedAvg aggregation strategy may be disproportionately influenced by dominant client-specific patterns and may not fully account for cross-client heterogeneity, which can affect both convergence stability and local predictive performance. To address this, we selected the model corresponding to the number of rounds that minimized the loss, just before the onset of instability. Future work could explore more advanced aggregation strategies beyond FedAvg or implement FL-specific early stopping mechanisms that monitor rounds rather than epochs to mitigate this issue.

The experiment shows that FL introduces negligible data overhead for the small model used here (approximately 200 KB per transfer); larger models would incur proportionally higher overhead. With 200 communication rounds, each client receives ~80 MB and uploads ~40 MB, while the server handles ~240 MB of uploads and ~120 MB of downloads. Given the ~200 KB model size per transfer, client-side communication costs remain minimal. These results confirm that FL enables efficient and privacy-preserving collaboration without excessive network overhead.

Conventional FL incurs a high network communication burden because the model parameters are fully exchanged between the aggregation server and clients as many times as the number of training rounds. Recent work has shown that serverless aggregation can greatly benefit these FL systems \cite{fedless, lambdafl}.

\subsection*{Limitations and Future Work}

Despite promising results, this study has several limitations. First, while our FL framework successfully demonstrated knowledge transfer between three regions with asynchronous outbreaks, no alternative aggregation strategy, personalization method, or sensitivity analysis was tested in this study, so the robustness of the reported performance to these choices remains to be established.. Second, our model primarily utilized temporal epidemic data; incorporating auxiliary variables such as intra-city mobility, weather conditions, or viral variant prevalence could further improve forecast robustness. Finally, future research should explore the integration of more advanced architectures, such as Temporal Fusion Transformers, within the federated environment to further push the boundaries of decentralized epidemic intelligence.

Several further limitations should be noted. The experiments used publicly available, aggregated case-count data that could, in principle, be centralized; the federated setting therefore served as a controlled testbed rather than a scenario in which centralization was impossible and the added value of FL should be confirmed in deployments involving genuinely restricted data. In addition, we did not compare the proposed models with conventional time-series approaches such as ARIMA or regression-based methods; consequently, the observed improvements over the local baseline cannot be attributed solely to federated learning, as they may partly reflect cross-region data pooling or the flexibility of the LSTM architecture itself. Moreover, the federated model was selected from the training round with the minimum loss before instability became pronounced, which may bias the reported results upward.

Another important limitation relates to the non-independent and identically distributed (non-IID) nature of the data across regions, reflecting differences in population dynamics, reporting practices, and public health interventions. This heterogeneity may contribute to training instability, as observed in the oscillations of the loss function during later federated learning rounds. Although our approach mitigates some of these effects through aggregation, more advanced strategies such as personalized federated learning or adaptive aggregation methods could further improve convergence and local performance. Exploring these approaches remains an important direction for future work.

Compared with localized or centralized non-federated approaches, FL introduces additional technical and computational requirements, including distributed coordination across sites, repeated communication rounds, and longer end-to-end training time. In our implementation, these overheads were acceptable for the relatively small LSTM model and limited number of participating clients, but may become more pronounced for larger models or broader deployments.

Future work will investigate transformer-based architectures, personalized federated learning strategies to address non-IID data, and additional analyses of model behavior in critical scenarios such as epidemic peak anticipation. We also plan to incorporate richer contextual features, including mobility and policy data, to improve robustness and generalization across regions.

\section*{Conclusions}

This study constructed and evaluated a proof-of-concept a cross-border FL system spanning multiple centers in Guangdong, Macau, and Portugal. The research findings demonstrate that the F-LSTM model outperforms standalone LSTM models in forecasting accuracy. Relative to the centralized benchmark, the federated model achieved comparable accuracy for 1-day-ahead forecasts but was less accurate for 7-day-ahead forecasts, particularly in Guangdong and Macau.

Specifically, our research shows that removing Portugal from the FL process severely degraded the model’s ability to forecast epidemic trends in Macau and Guangdong, regions with distinct demographic and epidemiological patterns. This clearly suggests that federated models can benefit from geographically and epidemiologically diverse data, even when raw data cannot be shared due to regulatory or ethical constraints.

In contrast to previous federated learning studies, we evaluate our approach using real, heterogeneous epidemic surveillance data from three different regions in a geographically distributed deployment with heterogeneous hardware. This approximates operational aspects faced by many international public health collaborations, such as distributed infrastructure, heterogeneous data distributions, and communication constraints, although the data themselves were public.

Moreover, we go beyond model performance by including a quantitative analysis of network overhead, communication load, and training time, addressing a frequent concern in the practical deployment of FL systems. This comprehensive assessment, rarely done in previous works, provides valuable evidence for future implementations of privacy-preserving AI in health.


Ultimately, this research contributes to an important challenge in global health: how to support collaborative epidemic forecasting across jurisdictions without compromising privacy or sovereignty. Our findings provide a concrete example of how FL can be deployed in this context, offering technical insights for privacy-preserving cross-border forecasting.

This work offers both technical and policy insights for the implementation of privacy-preserving AI in global health, serving as a reference implementation for future collaborations between jurisdictions.

\section*{Contributorship}
Ruibin Liu designed the experiment and contributed to the main text. Xuanfeng Li, Zige Liu contributed to the drafting of the manuscript. Francisco Fernandes and Andr\'e Salgado helped design and implement some of the experimental work and contributed to the writing of the document. Corresponding author Arlindo Oliveira spearheaded the development of the overarching architecture, meticulously designing the framework of the three-center system. Corresponding author Zifeng Yang, Chitin Hon and in particular Ana Freitas conceived the overarching FL technical architecture and critically reviewed the entire manuscript. All authors contributed to the reading and revising of the document.

\section*{Data Availability}
The source code of the L-LSTM, C-LSTM, and the proposed F-LSTM models, along with the dataset and forecasting results generated in this study are publicly available in our GitHub repository at: https://github.com/ruibinliu/flwr-experiment. This repository is designed to ensure the full reproducibility of our experimental findings and performance metrics.

\section*{Funding}
This work was supported by the National Key Research and Development Program of China (Grant No. 2024YFE0214800); the Science and Technology Development Fund of Macau SAR (Grant No. FDCT 0002/2024/RDP); national funds through the Funda\c c\~ ao para a Ci\^ encia e a Tecnologia in Portugal, with references DOI:10.54499/UIDB/50021/2020 and DOI:10.54499/DL57/\\2016/CP1368/CT0002; the Major Project of Guangzhou National Laboratory (Grant No. GZNL2024A01004); and the Self-Supporting Program of Guangzhou Laboratory (Grant No. SRPG22-007); the Science and Technology Development Program of Guangdong Province (2025B1212030002); Engineering Technology Research (Development) Center of Ordinary Colleges and Universities in Guangdong Province (Grant No. 2024GCZX010) and Guangdong Engineering Technology Research Center (Grant No. 2024A137).

\section*{Declaration of Competing Interest}
The authors declared no potential conflicts of interest with respect to the research, authorship, and/or publication of this article.

\section*{Ethical Approval}

This study utilized publicly available, aggregated epidemiological data from international repositories (WHO and China CDC), containing no identifiable individual-level information, and was therefore exempt from institutional review board (IRB) approval. All data used complied with the respective data governance frameworks of the participating regions. In particular, the study design adhered to the principles of the General Data Protection Regulation (GDPR) in the European context and corresponding data protection regulations in China and Macau SAR. No personal or sensitive data were transferred across jurisdictions, as the federated learning framework ensured that all data remained locally stored and only model parameters were exchanged.

\section*{Guarantor}
LR, the first author, takes full responsibility of this article.

\section*{Acknowledgements}
We thank the editor and the reviewers for their insightful comments and constructive suggestions, which have significantly improved the quality and clarity of the manuscript.

\bibliographystyle{unsrtnat}
\bibliography{references}

\end{document}